\documentclass[runningheads]{llncs}

\usepackage{eccv}

\usepackage{eccvabbrv}

\usepackage{graphicx}
\usepackage{booktabs}
\usepackage{bm}
\usepackage[accsupp]{axessibility}  %

\usepackage[pagebackref,breaklinks,colorlinks,citecolor=eccvblue]{hyperref}

\usepackage{orcidlink}

\begin{document}

\title{Backward-Compatible Aligned Representations via an Orthogonal Transformation Layer} 

\titlerunning{Orthogonal Compatible Aligned (OCA) Representations}

\author{Simone Ricci \orcidlink{0000-0001-9838-6076} \and
Niccolo Biondi \orcidlink{0000-0003-1153-1651} \and
Federico Pernici \orcidlink{0000-0001-7036-6655} \and
Alberto Del Bimbo \orcidlink{0000-0002-1052-8322}}

\authorrunning{S.~Ricci et al.}

\institute{DINFO (Department of Information Engineering), University of Florence, Italy, \\
MICC (Media Integration and Communication Center), \\
\email{\{\tt name\}.\{\tt surname\}@unifi.it}}
\maketitle

\begin{abstract}

Visual retrieval systems face significant challenges when updating models with improved representations due to misalignment between the old and new representations. The costly and resource-intensive backfilling process involves recalculating feature vectors for images in the gallery set whenever a new model is introduced. To address this, prior research has explored backward-compatible training methods that enable direct comparisons between new and old representations without backfilling.
Despite these advancements, achieving a balance between backward compatibility and the performance of independently trained models remains an open problem. 
In this paper, we address it by expanding the representation space with additional dimensions and learning an orthogonal transformation to achieve compatibility with old models and, at the same time, integrate new information. 
This transformation preserves the original feature space's geometry, ensuring that our model aligns with previous versions while also learning new data. 
Our Orthogonal Compatible Aligned (OCA) approach eliminates the need for re-indexing during model updates and ensures that features can be compared directly across different model updates without additional mapping functions.
Experimental results on CIFAR-100 and ImageNet-1k demonstrate that our method not only maintains compatibility with previous models but also achieves state-of-the-art accuracy, outperforming several existing methods.
Code at: \href{https://github.com/z3n0e/OCA}{GitHub repository}.

\keywords{Deep Learning \and Representation Learning \and Compatible Representation Learning \and Orthogonal Transformation}
\end{abstract}

\section{Introduction}

\label{sec:intro}

Visual retrieval systems operate by matching images from a stored dataset (the gallery set) to input images (the query set). 
This process involves using a trained representation model to encode all gallery images into feature representations. 
When queries are available, the system retrieves the most similar gallery representations.

With advancements in the expressive power of representation models, updating the gallery with newer models is necessary to obtain improved performance \cite{shen2020towards}.
This is particularly challenging when the new model is trained independently of the old one, or they have different network architectures, resulting in completely different and incompatible representations. 
Consequently, recomputing the feature vectors for all images in the gallery set, a process known as backfilling or re-indexing, becomes essential. 
However, this can be prohibitively expensive or even infeasible for real-world galleries containing vast numbers of images.

Recent research addresses the challenge of avoiding the backfilling of the gallery set by learning model representations that can be directly compared without reprocessing the gallery data with an improved model. 
These representations are referred to as compatible \cite{shen2020towards, zhang2021hot, ramanujan2022forward, meng2021learning, biondi2023cores, zhou2023bt, biondi2024stationary}. 
The seminal work by \cite{shen2020towards} proposed learning compatible representations using an influence loss into the training objective of the new model, which aligns the new representation with the previous one. 
However, training the new model using this loss reduces its performance compared to training the same model independently \cite{ramanujan2022forward}. 
To overcome this issue, subsequent studies \cite{zhang2021hot, meng2021learning, zhang2022towards} have proposed different loss functions, but these efforts have met with limited success.
In another line of research, \cite{wang2020unified, ramanujan2022forward} have explored learning a lightweight transformation between old and new model representations, aiming to fully leverage the improvements provided by the independently trained version of the new model. 
However, learning these transformations still demands additional training time and a resource-intensive process of mapping all the data in the gallery with these transformation functions. 
More recently, the conflict between backward compatibility and new model performance has been addressed by \cite{zhou2023bt}. This approach expands the representation space with additional dimensions, allowing the old portion of the feature space to align with the old model while incorporating new knowledge using the independently trained version of the new model in the remaining feature dimensions. In their method, the backbone generates a representation optimized to align with a newly trained independent model's representation through a matching and classification loss. A subset of this representation undergoes a learnable basis transformation, preserving information from the new representation. The new representation is then projected into a compact embedding space and merged with part of the transformed new representation. A second basis transformation is applied to this merged space to match the old model's representation. The part of the transformed new representation that is not merged captures additional information that may be incompatible with the old model but is useful for improving representation quality.

\begin{figure*}[t]
  \centering
   \includegraphics[width=0.77\linewidth]{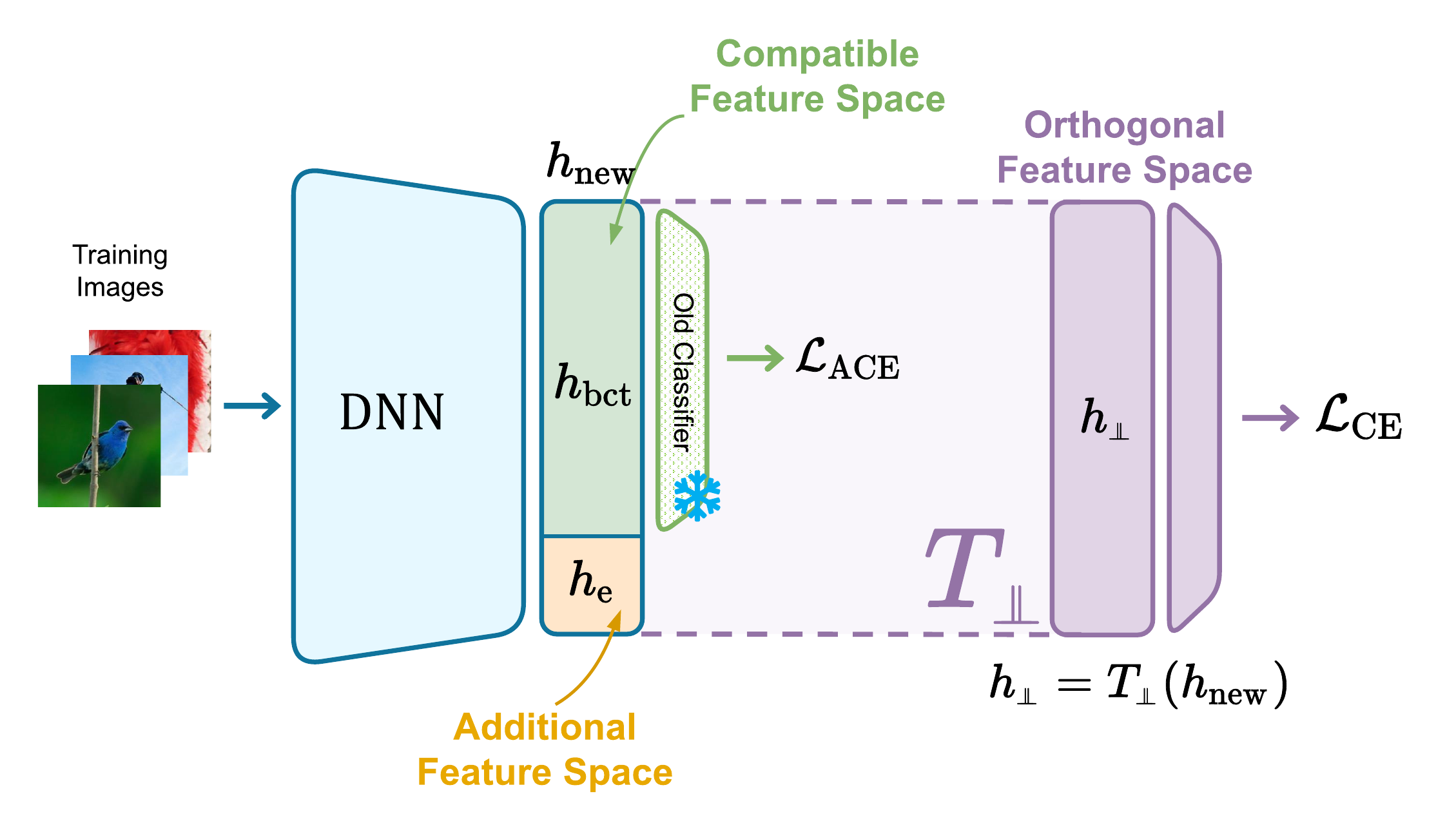}
   \caption{\textbf{Overview of our method.} The DNN backbone generates representations in a feature space $h_{\rm new}$.
   This feature space is divided into two different parts: $h_{\rm btc}$ is the learned compatible representation space according to $\mathcal{L}_{\rm ACE}$, while $h_{\rm e}$ is an extra feature space used to learn new information from new data without negatively affecting the old feature space configuration.
   \mbox{$h_{\rm new} = [h_{\rm bct}|h_{\rm e}]$} is then transformed with $ T_{\scriptscriptstyle{\pmb{\perp}}}$ into $h_{\scriptscriptstyle{\pmb{\perp}}}$ and then used for classification using $\mathcal{L}_{\rm CE}$.}
   \label{fig:method_overview}
\end{figure*}

In this paper, we address the dual challenges of learning backward-compatible representations while maintaining performance comparable to the independently trained version of the new model. To achieve this, we expand the feature space of the new model by adding extra dimensions relative to the old model. New information can be integrated into this representation space without necessitating a new independently trained model by learning an orthogonal transformation function (Fig. \ref{fig:method_overview}). This orthogonal transformation preserves the geometry of the compatible learned feature space, ensuring that the new model aligns with the old one. Meanwhile, in the additional dimensions not affected by the compatible training, the model incorporates new information. At inference time, the orthogonal transformation is discarded, and visual search is performed with features extracted from the original feature space prior to the transformation. This approach ensures that features generated with our method can be directly compared across multiple model updates without the need for composing mapping functions when comparing non-sequential representations.

Our experimental results on CIFAR-100 and ImageNet-1k show that, using the proposed approach, the new representation is compatible with the old one while achieving the state-of-the-art accuracy against several compared methods.

In summary, the contributions of this paper are:
\begin{itemize}
\item We introduce a method that expands the feature space of models to allow the integration of new information without losing backward compatibility or degrading performance.
\item We employ an orthogonal transformation that preserves the geometry of the original feature space, ensuring that new models align with older versions for consistent results.
\item We demonstrate the effectiveness of our approach through experiments on \mbox{CIFAR-100} and ImageNet-1k, achieving state-of-the-art accuracy and proving compatibility across model updates.
\end{itemize}

\section{Related Works}
\label{sec:related}

Compatible training aims to learn representations that can be used interchangeably when updating a model.
The objective is to establish a unified representation space where it is possible to directly compare representations from various models. 
Methods in this area can be categorized into \textit{direct} comparison between old and new representations~\cite{shen2020towards, zhang2022towards, zhang2021hot} or mapping-based approaches~\cite{Chen_2019_CVPR, wang2020unified, ramanujan2021forward, meng2021learning}. 
Direct compatibility can be obtained typically involving the usage of auxiliary loss functions. 
This is introduced to align representation with the old classifier prototypes~\cite{shen2020towards} or to refine the prototype neighbor structure with a fully-connected graph~\cite{zhang2022towards}.
Mapping-based methods differ in how they learn the transformation function used to compare updated and old representations. 
In particular, \cite{meng2021learning} also allows for direct comparison by simply imposing the transformation module to be the identity. 

Nonetheless, certain notable drawbacks associated with these methodologies persist. 
Relying on an auxiliary loss hinders the new model's ability to achieve comparable performance with the independently trained version of the new model, while the mapping-based approaches require additional training to learn the mapping function after the training of the new models and then a lightweight backfilling process \cite{ramanujan2022forward} to extract compute the mapping of the existing gallery features. 
In addition to this, when the model is subjected to multiple updates, to compare non-sequential representations the solution is to concatenate multiple mapping functions, which again increases the re-processing cost. 
The research in \cite{zhou2023bt} shows that a trade-off exists within the concept of backward-compatibility, as delineated in \cite{shen2020towards}, and the performance that the new model has. 
To address this issue, they expand the feature space of the new model to retain new information in an extra space while aligning it to the old one. 
This work has connections with \cite{biondi2023cores,biondi2024stationary}. 
This work theoretically and empirically showed that the stationarity of the representations (i.e., remaining aligned across several model updates) is crucial to achieving compatibility. 
To achieve stationarity, the feature space of the model is pre-allocated since the beginning of training according to a simplex configuration  \cite{pernici2021regular} to accommodate current data while keeping some free space for future classes. 

Several other works have delineated new definitions of compatibility~\cite{budnik2020asymmetric,duggal2021compatibility,yan2021positive,zhang2021hot,seo2023online} or studied compatibility under a continual learning scenario \cite{biondi2023cl2r, iscen2020memory, Wan_2022_CVPR, Cui_2024_CVPR}.
In this paper, we follow the definition given by~\cite{shen2020towards} and we update the model retraining from scratch the model every time new data is available using also the whole old data, i.e., therefore avoiding the catastrophic forgetting issue \cite{robins1993catastrophic, mccloskey1989catastrophic}.

\section{Methodology}

\subsection{Backward-Compatible Training}
Let $\phi_{\rm old}$ be the initial representation model trained using cross-entropy loss on an initial labeled training set $\mathcal{D}_{\rm old} = \{\mathbf{x}_i, y_i \}_{i=1}^{C_{\rm old}}$, where $\mathbf{x}_i$ is a generic image with label $y_i$. After training, $\phi_{\rm old}$ is used to extract features from a gallery $\mathcal{G}=\{\mathbf{x}_i, y_i \}_{i=1}^{C}$ and a query set $\mathcal{Q}=\{\mathbf{x}_i, y_i \}_{i=1}^{C}$, with $C$ being an arbitrary number of classes. In the following, we refer to the set of features from the gallery and the query set obtained with $\phi_{\rm old}$ as $\phi^{\mathcal{G}}_{\rm old}$ and $\phi^{\mathcal{Q}}_{\rm old}$, respectively. When a new set of images $\mathcal{X}$ becomes available, $\phi_{\rm new}$ is trained using $\mathcal{D}_{\rm new} = \mathcal{D}_{\rm old} \cup \mathcal{X}$. The newly added data $\mathcal{X}$ can have a similar distribution to the previous data $\mathcal{D}_{\rm old}$ or a completely different one. In this paper, we assume the worst-case scenario where $\mathcal{X}$ belongs to a different and non-overlapping set of classes than $\mathcal{D}_{\rm old}$, such that $\mathcal{D}_{\rm new} = \{\mathbf{x}_i, y_i \}_{i=1}^{C_{\rm new}}$.
Backward-compatible training aims to learn $\phi_{\rm new}$ in a way that allows direct comparison of the features of the query set extracted with the new model $\phi^{\mathcal{Q}}_{\rm new}$ with the features of the gallery set obtained with the old model $\phi^{\mathcal{G}}_{\rm old}$, thus avoiding the need to reprocess the gallery set with the new model $\phi^{\mathcal{G}}_{\rm new}$.

According to \cite{shen2020towards,wang2020unified}, the following definition of compatibility between representation models holds:

\begin{defn}[Backward Compatibility \cite{shen2020towards}] \label{def:compatibility-shen}
Two representation models $\phi_{\rm old}$ and $\phi_{\rm new}$ are compatible if $\forall \ \mathbf{x}_i , \mathbf{x}_j$ samples from the distribution of interest, with $i \neq j$, holds that:
\begin{subequations}\label{eq:compatible_set_dist}
\begin{align}
 \dist \big(\phi_{\rm old}(\mathbf{x}_i), \phi_{\rm new}(\mathbf{x}_j) \big) &\leq {\dist} \big(\phi_{\rm old}(\mathbf{x}_i), \phi_{\rm old}(\mathbf{x}_j) \big)  \label{eq:first} \\
  \text{ \rm with } y_i = y_j  \notag \\
  \dist \big(\phi_{\rm old}(\mathbf{x}_i), \phi_{\rm new}(\mathbf{x}_j) \big) &\geq \dist \big(\phi_{\rm old}(\mathbf{x}_i), \phi_{\rm old}(\mathbf{x}_j) \big) \label{eq:second}   \\
  \text{ \rm with } y_i \neq y_j  \notag
\end{align}
\end{subequations}
where $y_i, y_j$ are the corresponding labels of $\mathbf{x}_i, \mathbf{x}_j$ and $\dist(\cdot, \cdot)$ is a distance function.
\end{defn}
An intuitive interpretation of Def. \ref{def:compatibility-shen} is that the new model should perform as well as, or better than, the old model in grouping data of the same class. This implies that the distances between new and old feature points of the same class should be less than or equal to those between the old feature vectors (Eq. \ref{eq:first}). 
Simultaneously, the new model should be better or at least equal to the old model in discriminating data belonging to different classes. 
 Consequently, distances between new and old feature points of different classes should be greater than or equal to those between the old feature vectors (Eq. \ref{eq:second}).

However, it is important to note that Def. \ref{def:compatibility-shen} is not practical for real-world applications as it requires evaluating the pairwise distances between all data points. 
This requirement makes the criterion computationally intensive and challenging to implement at scale.
Therefore, as suggested by \cite{shen2020towards}, Def. \ref{def:compatibility-shen} is relaxed into the following Empirical Compatibility Criterion:
\begin{eqnarray} \label{eq:multistepecc}
M \big( \phi_{\rm new}^\mathcal{Q}, \phi_{\rm old}^\mathcal{G} \big) > 
M \big( \phi_{\rm old}^\mathcal{Q}, \phi_{\rm old}^\mathcal{G} \big)
\end{eqnarray}
where $M$ is a performance metric.

\subsection{Backward-Compatibility via Representations Alignment and Orthogonal Transformation}
\label{sec:merge}

In this section, we present how we train the new model $\phi_{\rm new}$ to learn a representation that is backward-compatible with $\phi_{\rm old}$ according to Def.~\ref{def:compatibility-shen}, while improving the discrimination capability of the new model using the new incoming data. 
Fig. \ref{fig:method_overview} shows an overview of our approach that is motivated in the following. 

Theoretical \cite{biondi2024stationary} and empirical investigations \cite{shen2020towards, wang2020unified, meng2021learning, biondi2023cl2r, biondi2023cores} have shown that to achieve compatibility between $\phi_{\rm new}$ and $\phi_{\rm old}$, the new representation space should be aligned as closely as possible to the old one.
To this end, we align the representation space of $\phi_{\rm new}$ with the old one by minimizing the distance between the features and the class prototypes of the old classifier, which is kept fixed during the learning of the new model. An old class prototype $W_{\rm old}^{(y)}$ is obtained by averaging all features extracted from the old network for each image of the class $y$.
In particular, we optimize the influence loss via cross-entropy loss $\mathcal{L}_{\textsc{ce}}(W_{\rm old}, \phi_{\rm new}(\mathbf{x}))$ between features extracted with the new model $\phi_{\rm new}$ and the old classifier prototypes $W_{\rm old}$ as in \cite{shen2020towards}. Motivated by \cite{yang2022we}, which shows that cross-entropy optimization does not achieve optimal alignment between learnable features and fixed class prototypes, we enforce feature stationarity by optimizing the cosine distance ($\mathcal{L}_{\angle}$) between the newly learned representations and their corresponding fixed old classifier prototypes, thereby directly enhancing the alignment property of the cross-entropy loss.
To achieve backward-compatibility, we optimize the following Aligning Compatible Embedding (ACE) loss:
\begin{equation}
    \mathcal{L}_{\textsc{ace}} = \lambda_{1} \cdot  \mathcal{L}_{\textsc{ce}}(W_{\rm old}, \phi_{\rm new}(\mathbf{x})) + \lambda_{2} \cdot  \mathcal{L}_{\angle}
\end{equation}
where $\lambda_{1}$ and $\lambda_{2}$ are two weighting factors, $\mathcal{L}_{\textsc{ce}}$ is the cross-entropy loss defined as

\begin{equation} \label{eq1}
\mathcal{L}_{\textsc{ce}} = - \sum\limits_{(\mathbf{x}, y) \in B} \sum_{i=1}^{C_{\rm new}} y_i \log \left( \frac{\exp(\phi_{\rm new}(\mathbf{x}) \cdot W_{\rm old}^{(i)})}{\sum_{j=1}^{C_{\rm new}} \exp(\phi_{\rm new}(\mathbf{x}) \cdot W_{\rm old}^{(j)})} \right)
\end{equation}

and $\mathcal{L}_{\angle}$ is defined as
\begin{equation} \label{eq2}
\mathcal{L}_{\angle} = - \sum\limits_{(\mathbf{x}, y) \in B} \left(1 - \frac{\phi_{\rm new}(\mathbf{x}) \cdot W_{\rm old}^{(y)}}{\lVert \phi_{\rm new}(\mathbf{x}) \lVert \lVert W_{\rm old}^{(y)}\lVert }\right)
\end{equation}
on the mini-batch $B$.
However, there is an inherent trade-off \cite{zhou2023bt} between training backward-compatible representations, which avoid the need for backfilling, and achieving the performance of a model trained directly on $\mathcal{D}_{\rm new}$.
This is because the new backward-compatible feature space $h_{\rm bct}$ is constrained to align with the old one, which prevents the model from learning a new feature space structure that could accommodate all the new information from the latest data.
Due to this, there is a drop in performance as the new model cannot properly assimilate the new knowledge from the incoming data. 
To address this issue while preserving the compatibility of the representations, we extend the feature space with extra dimensions $h_{\rm e}$ alongside the aligned compatible embeddings $h_{\rm bct}$. 
These additional dimensions $h_{\rm e}$ are not optimized by the ACE loss since they are used to accommodate new data.

As demonstrated by \cite{zhou2023bt}, directly training the new model on the expanded feature space \mbox{$h_{\rm new} = [h_{\rm bct}|h_{\rm e}]$} can lead to incompatible representations. 
This is because training in a higher-dimensional space tends to alter the geometric structure of the previous space, even if $h_{\rm bct}$ is constrained to align with the old one. 
Consequently, we apply an orthogonal transformation to $h_{\rm new}$, resulting in $h_{\scriptscriptstyle{\pmb{\perp}}}$, which lies in a representation space with the same dimensionality and geometric configuration as $h_{\rm new}$.
This transformation is achieved through a learnable orthogonal transformation layer $ T_{\scriptscriptstyle{\pmb{\perp}}}$, which is used to obtain the new orthogonal feature space $h_{\scriptscriptstyle{\pmb{\perp}}} = T_{\scriptscriptstyle{\pmb{\perp}}}(h_{\rm new})$. 
We define the weight of the linear transformation $T_{\scriptscriptstyle{\pmb{\perp}}}$ as $Q$, making $T_{\scriptscriptstyle{\pmb{\perp}}}$ a learnable orthogonal transformation that ensures $T_{\scriptscriptstyle{\pmb{\perp}}}^{\top} T_{\scriptscriptstyle{\pmb{\perp}}}^{\vphantom{\top}} = I$, where $I$ is as the identity matrix. 
To constraint a fully-connected layer to learn such a transformation, we use a matrix $A$, where $A$ is any skew-symmetric matrix (so that $A^{\top} = -A$) with learnable parameters randomly initialized. 
The orthogonal matrix $Q$ is then obtained using the exponential map $Q = e^A$.
Applying the orthogonal transformation $T_{\scriptscriptstyle{\pmb{\perp}}}$ to any representation preserves all geometrical information and maintains the quality of the representations, as it holds that:
\begin{align*}
    h_i^\top h_j =  T_{\scriptscriptstyle{\pmb{\perp}}}(h_i)^\top  T_{\scriptscriptstyle{\pmb{\perp}}}(h_j)
\end{align*}
where $h_i$ and $h_j$ are two generic learned representations.

The transformed embeddings $h_{\scriptscriptstyle{\pmb{\perp}}}$ are finally optimized through the learned classifier $W$ to learn incoming information from the new data.
The orthogonal transformation, due to its orthogonal column constraint imposed by $T_{\scriptscriptstyle{\pmb{\perp}}}^{\top} T_{\scriptscriptstyle{\pmb{\perp}}}^{\vphantom{\top}} = I$, ensures that the angles and norms of the input space are preserved, thereby maintaining the previous geometry in the learned feature space $h_{\rm new}$, especially for the compatible learned subspace $h_{\rm bct}$. This constrains the model to learn new information in the additional space $h_{\rm e}$ without disrupting the geometric structure of $h_{\rm bct}$. The orthogonal transformation layer ensures that the geometry of the compatible learned feature space $h_{\rm bct}$ remains unchanged. The cross-entropy loss, computed using the new classifier $W$, refines the new knowledge in the extra embedding dimensions $h_{\rm e}$.

The overall loss used in our approach is thus:
\begin{equation}
    \mathcal{L} =  \mathcal{L}_{\textsc{ce}}(W, \phi_{\rm new}(\mathbf{x})) + \mathcal{L}_{\textsc{ace}}
\end{equation}

\begin{figure*}[t]
  \centering
   \includegraphics[width=0.9\linewidth]{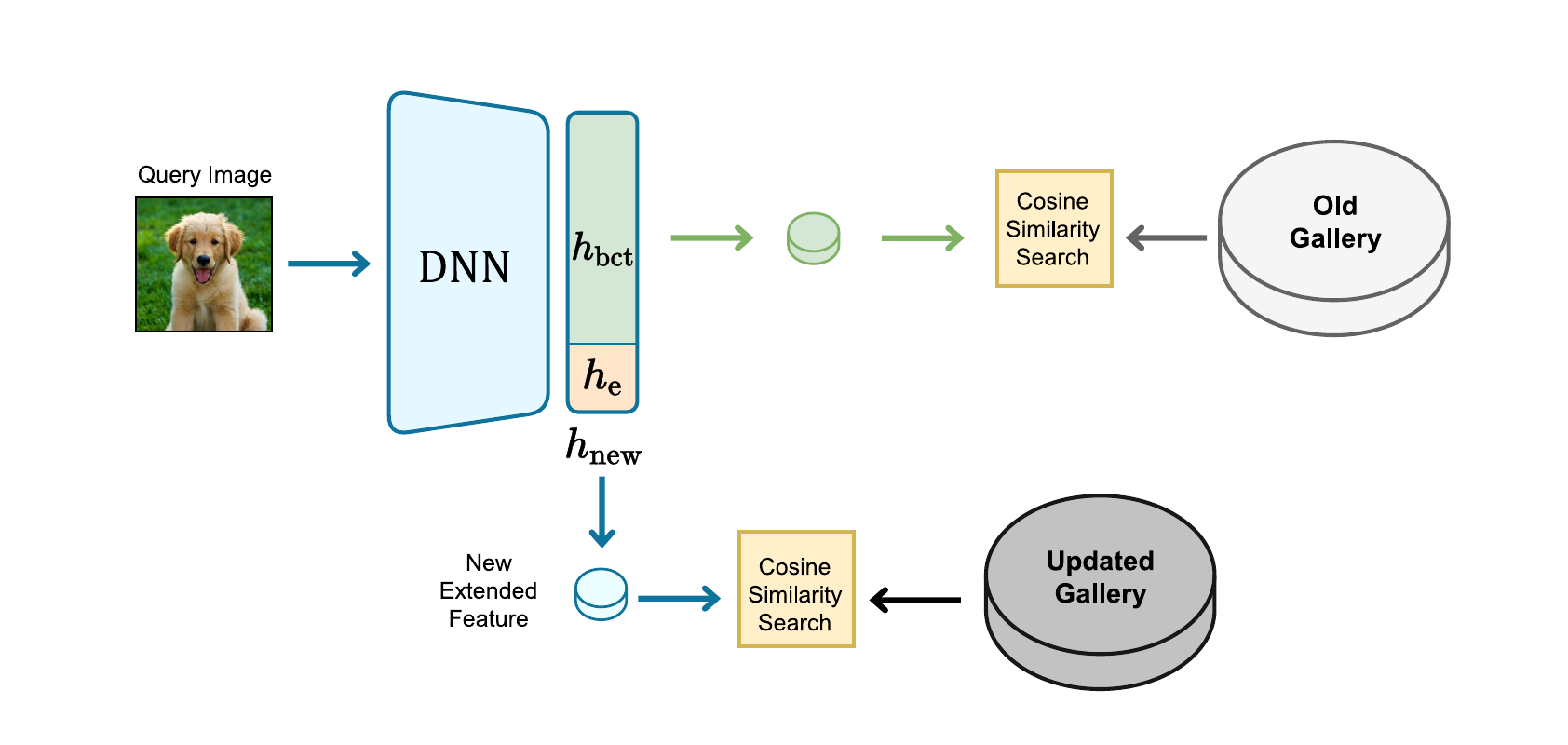}
   \caption{\textbf{Overview of our method at inference time.} The DNN backbone model produces representations within a feature space $h_{\rm new}$. This space is divided into two parts: $h_{\rm btc}$ is the compatible representation space. Its representations are used to perform visual search directly with the old gallery features without using the orthogonal transformation function that we discard after training.
   Representations \mbox{$h_{\rm new} = [h_{\rm bct}|h_{\rm e}]$} are instead used to match with the updated gallery to be as close as possible to the performance of the independently trained version of the new model.}
   \label{fig:method_inference}
\end{figure*}

It is worth noticing that, after learning, we use only $h_{\rm new}$ at inference time to perform image search/retrieval as shown in Fig. \ref{fig:method_inference}, while $ T_{\scriptscriptstyle{\pmb{\perp}}}$, $h_{\scriptscriptstyle{\pmb{\perp}}}$, and the new classifier $W$ are discarded as they are used only to learn additional knowledge from the new data $\mathcal{D}_{\rm new}$.

\section{Experimental Results}
\label{sec:experiments}

In this section, we present our experimental results that (1) evaluate our approach alongside established backward compatible representation learning techniques according to the criteria outlined in Eq.~\ref{eq:multistepecc}, (2) examine our approach's capacity to address incremental scenarios in data, or what \cite{shen2020towards} identified as open classes (for example, where the previous model was trained on 500 ImageNet classes while the updated model uses 1000 ImageNet classes)

\subsection{Datasets}
In this paper, we make use of the following datasets:
\begin{itemize}
    \item \textbf{CIFAR-100}\cite{Krizhevsky2009LearningML}: It is a small image classification dataset of 100 classes divided into 50000 images of the training set and 10000 of the test set. We will refer to CIFAR-50 as the partition consisting of all the samples from the first 50 classes.
    
    \item \textbf{ImageNet-1k} \cite{deng2009ImageNet}: It is a large-scale image recognition dataset proposed for the ILSVRC 2012 challenge. It has 1000 image classes with about 1k images per class. We follow the same partitioning as in \cite{ramanujan2022forward}. We consider the images from the first 500 classes as ImageNet-500.
\end{itemize}

\subsection{Evaluation Metrics}

\noindent \textbf{Mean Average Precision (mAP):} The mAP is a standard metric in compatibility, where precision and recall evaluations are summarized by calculating the area under the precision-recall curve. The average precision within the recall interval [0.0, 1.0] is computed.

\noindent \textbf{Cumulative Matching Characteristics (CMC):} CMC refers to the top-k accuracy, where gallery representations are ranked based on their similarity to the query representation. A match is correct if a representation of the same class appears within the first k gallery entries. We report CMC-1 (top-1 accuracy) for all models.

\noindent We construct a distance matrix between all the representations obtained from the query and gallery sets under consideration. We use the cosine similarity between two feature vectors as the measure of distance. On top of this matrix, we evaluate the mAP and CMC-1 metrics. During the retrieval process, for a method that utilizes additional dimensions, it is relatively simple to differentiate between the gallery samples featuring the old representations and those displaying the concatenated representations, thanks to the difference in size. We zero-pad the old representations stored in the gallery during comparison with the new representations.

\subsection{Compared Methods}
We compare our method OCA against the following approaches:

\begin{itemize}
    \item \textbf{Independent}: $\phi^{\rm I}_{\rm new}$ is trained from scratch with new data, without taking into consideration any backward compatibility method.
    
    \item \textbf{BCT} \cite{shen2020towards}: It is a widely adopted baseline in recent studies \cite{ramanujan2022forward}. BCT employs a classification loss regularized by an "influence loss" during the training process to ensure backward compatibility. In BCT, denoted $W$ as the trainable classifier of $\phi^{\rm BTC}_{\rm new}$ , and $W_{\rm old}$ as the fixed old classifier obtained by the old representation $\phi_{\rm old}$, the loss function comprises two terms:
    \begin{align*}
    \mathcal{L}_{\rm BCT}(\phi, W, \mathbf{x}) = \mathcal{L}_{\textsc{ce}}(W, \phi|\mathbf{x}) + \lambda \mathcal{L}_{\textsc{ce}}(W_{\rm old}, \phi|\mathbf{x}),
    \end{align*}
    where $\lambda$ is a hyperparameter that weights the influence loss.
    
    \item \textbf{BT$^2$} \cite{zhou2023bt}: It employs an embedding dimension expansion of 32, utilizing a combination of cosine similarity loss and BCT influence loss for matching $\phi_{\rm old}$. Additionally, it enhances the learned representation by matching $\phi^{\rm I}_{\rm new}$ with an additional cosine similarity loss. The supplementary feature space is learned through two trainable changes of basis to prevent the introduction of information that may disrupt the compatible learned information.
\end{itemize}

\subsection{Implementations Details}
All baselines and our method employ ResNet50 models with an initial embedding size of 128 as backbone, trained using the Adam optimizer, with a learning rate of 0.001 and a batch size of 128 over 100 epochs. For our method, we expand the embedding dimension by 32 and set $\lambda_{1}=10$ and $\lambda_{2}=5$.

\subsection{Experimental Results}

\begin{table}[t]
\centering
\caption{Results of the mean Average Precision (mAP) and Cumulative Matching Characteristics (CMC) metrics from trials carried out on CIFAR-50 and CIFAR-100 datasets. These experiments utilized the Resnet50-128 architecture in both old and new models. The $/$ symbol distinguishes the model that processes the query (left of $/$) from the model that processes the gallery set (right of $/$).}
\label{table:CIFAR-100}
\setlength{\tabcolsep}{8pt}
\begin{tabular}{lccc}
\toprule[1pt]
\textbf{Method} & \textbf{Case} & \textbf{mAP@1.0} & \textbf{CMC-1} \\
\toprule[1pt]
Initial Model & $\phi_{\rm old}/\phi_{\rm old}$ & 23.32 &  31.32 \\
\hline
\multirow{2}{*}{Independent} & $\phi_{\rm new}^{\rm I}/\phi_{\rm old}$ & 01.29 & 01.02 \\
 & $\phi_{\rm new}^{\rm I}/\phi_{\rm new}^{\rm I}$ & 45.35 & 56.75 \\
\hline
\multirow{2}{*}{BCT \cite{shen2020towards}} & $\phi_{\rm new}^{\rm BCT}/\phi_{\rm old}$ & 25.14 & 36.73 \\
 & $\phi_{\rm new}^{\rm BCT}/\phi_{\rm new}^{\rm BCT}$ & 43.89 & 54.76 \\
\hline
\multirow{2}{*}{BT$^2$ \cite{zhou2023bt}} & $\phi_{\rm new}^{\rm BT^2}/\phi_{\rm old}$ & 26.05 & 38.64 \\
 & $\phi_{\rm new}^{\rm BT^2}/\phi_{\rm new}^{\rm BT^2}$ & 50.36 & 61.77 \\
\hline
\multirow{2}{*}{\textbf{OCA}} & $\phi_{\rm new}^{\rm OCA}/\phi_{\rm old}$ & \textbf{26.35} & \textbf{41.37} \\
 & $\phi_{\rm new}^{\rm OCA}/\phi_{\rm new}^{\rm OCA}$ & \textbf{52.06} & \textbf{62.02} \\
\bottomrule[1pt]
\end{tabular}
\end{table}

\begin{table}[t]
\centering
\caption{Results of the mean Average Precision (mAP) and Cumulative Match Characteristic (CMC) metrics from trials carried out on ImageNet-500 and ImageNet-1k datasets. These experiments utilized the Resnet50-128 architecture in both old and new models. The $/$ symbol distinguishes the model that processes the gallery (left of $/$) from the model that processes the query set (right of $/$).}
\setlength{\tabcolsep}{8pt}
\begin{tabular}{lccccc}
\toprule[1pt]
\textbf{Method} & \textbf{Case} & \textbf{mAP@1.0} & \textbf{CMC-1}\\
\toprule[1pt]
Initial Model & $\phi_{\rm old}/\phi_{\rm old}$ & 32.33 & 43.53  \\
\hline
\multirow{2}{*}{Independent} & $\phi_{\rm new}^{\rm I}/\phi_{\rm old}$ & 00.14 & 00.12 \\
 & $\phi_{\rm new}^{\rm I}/\phi_{\rm new}^{\rm I}$ & 55.53 & 69.11 \\
\hline
\multirow{2}{*}{BCT \cite{shen2020towards}} & $\phi_{\rm new}^{\rm BCT}/\phi_{\rm old}$ & 35.81 & 48.56 \\
 & $\phi_{\rm new}^{\rm BCT}/\phi_{\rm new}^{\rm BCT}$ & 54.44 & 67.57  \\
\hline
\multirow{2}{*}{BT$^2$ \cite{zhou2023bt}} & $\phi_{\rm new}^{\rm BT^2}/\phi_{\rm old}$ & 36.55 & 50.21  \\
 & $\phi_{\rm new}^{\rm BT^2}/\phi_{\rm new}^{\rm BT^2}$ & 55.65 & 67.75  \\
\hline
\multirow{2}{*}{\textbf{OCA}} & $\phi_{\rm new}^{\rm OCA}/\phi_{\rm old}$ & \textbf{36.71} & \textbf{50.73}  \\
 & $\phi_{\rm new}^{\rm OCA}/\phi_{\rm new}^{\rm OCA}$ & \textbf{56.82} & \textbf{69.73} \\
\bottomrule[1pt]
\end{tabular}
\label{table:ImageNet}
\end{table}

\noindent \textbf{CIFAR-50 to CIFAR-100.} For this experiment, the $\phi_{\rm old}$ model is trained using the CIFAR-50 dataset, while the $\phi_{\rm new}$ model is trained on the complete CIFAR-100 dataset. We conduct retrieval tasks for metric evaluation on the CIFAR-100 validation set, which serves as both the gallery and query sets.

\noindent \textbf{ImageNet-500 to ImageNet-1k.} The initial model $\phi_{\rm old}$ is trained with the ImageNet-500 dataset, followed by training of $\phi_{\rm new}$ using the complete ImageNet-1k dataset. In our retrieval process evaluation, we use the ImageNet-1k validation set for both gallery and query purposes.

The results are shown in Table \ref{table:CIFAR-100} and \ref{table:ImageNet} for CIFAR-100 and ImageNet-1k, respectively.
We observe that Independent training is the only method that fails to achieve compatibility with the gallery extracted by $\phi_{\rm old}$ because it does not implement any compatible learning strategy. In contrast, BCT achieves compatibility of $\phi^{\rm BCT}_{\rm new}$ with the old gallery representations, but it shows a reduction in both performance metrics compared to Independent training when evaluating new query representations against new gallery representations, which aligns with the findings in \cite{zhou2023bt}.
BT$^2$ improves compatibility performance compared to BCT, thanks to an additional embedding dimension and the increased number of parameters provided by the two orthogonal matrices for the basis changes and a previously trained Independent model. Our method achieves the best results on both datasets, demonstrating its potential. Compared to BT$^2$, we add fewer parameters to the model because our Orthogonal Transformation layer and the auxiliary classifier are used only during training and are then completely removed. This allows us to leverage all the new information without hurting compatibility and increases model generalization, achieving also better performance than an Independent model when evaluating new query representations against new gallery representations.

\subsection{Ablation Studies}

\begin{table}[t]
\centering
\caption{Ablation on the influence of the extra space $h_{\rm e}$ dimentionality to the compatible training. Results of the mean Average Precision (mAP) and Cumulative Match Characteristic (CMC) metrics from trials carried out on CIFAR-50 and CIFAR-100 datasets. These experiments utilized the Resnet50-128 architecture in both old and new models. The $/$ symbol distinguishes the model that processes the gallery (left of $/$) from the model that processes the query set (right of $/$).}
\label{table:ablation1}
\setlength{\tabcolsep}{8pt}
\begin{tabular}{lccccc}
\toprule[1pt]
\textbf{Ext. Dim.} & \textbf{Method} & \textbf{Case} & \textbf{mAP@1.0} & \textbf{CMC-1} \\
\toprule[1pt]
\multirow{4}{*}{ } & \multirow{1}{*}{Initial Model} & $\phi_{\rm old}/\phi_{\rm old}$ & 23.32 &  31.32 \\
\hline
\multirow{4}{*}{ } & \multirow{2}{*}{Independent} & $\phi_{\rm new}^{\rm I}/\phi_{\rm old}$ & 01.29 & 01.02 \\
 &  & $\phi_{\rm new}^{\rm I}/\phi_{\rm new}^{\rm I}$ & 45.35 & 56.75 \\
\hline
\multirow{4}{*}{+1} & \multirow{2}{*}{BT$^2$ \cite{zhou2023bt}} & $\phi_{\rm new}^{\rm BT^2}/\phi_{\rm old}$ & 25.58 & 34.79 \\
 &  & $\phi_{\rm new}^{\rm BT^2}/\phi_{\rm new}^{\rm BT^2}$ & 44.42 & 59.84 \\
\cline{2-5}
 & \multirow{2}{*}{OCA} & $\phi_{\rm new}^{\rm OCA}/\phi_{\rm old}$ & 27.09 & 43.95 \\
 &  & $\phi_{\rm new}^{\rm OCA}/\phi_{\rm new}^{\rm OCA}$ & 50.87 & 61.04 \\
\hline
\multirow{4}{*}{+32} & \multirow{2}{*}{BT$^2$ \cite{zhou2023bt}} & $\phi_{\rm new}^{\rm BT^2}/\phi_{\rm old}$ & 26.05 & 38.64 \\
 &  & $\phi_{\rm new}^{\rm BT^2}/\phi_{\rm new}^{\rm BT^2}$ & 50.36 & 61.77 \\
\cline{2-5}
 & \multirow{2}{*}{OCA} & $\phi_{\rm new}^{\rm OCA}/\phi_{\rm old}$ & 26.35 & 41.37 \\
 &  & $\phi_{\rm new}^{\rm OCA}/\phi_{\rm new}^{\rm OCA}$ & 52.06 & 62.02 \\
 \hline
 \multirow{4}{*}{+64} & \multirow{2}{*}{BT$^2$ \cite{zhou2023bt}} & $\phi_{\rm new}^{\rm BT^2}/\phi_{\rm old}$ & 22.60 & 24.17 \\
 &  & $\phi_{\rm new}^{\rm BT^2}/\phi_{\rm new}^{\rm BT^2}$ & 50.36 & 62.87 \\
\cline{2-5}
 & \multirow{2}{*}{OCA} & $\phi_{\rm new}^{\rm OCA}/\phi_{\rm old}$ & 26.76 & 42.26 \\
 &  & $\phi_{\rm new}^{\rm OCA}/\phi_{\rm new}^{\rm OCA}$ & 51.69 & 61.03 \\
 \hline
 \multirow{4}{*}{+128} & \multirow{2}{*}{BT$^2$ \cite{zhou2023bt}} & $\phi_{\rm new}^{\rm BT^2}/\phi_{\rm old}$ & 12.43 & 08.98 \\
 &  & $\phi_{\rm new}^{\rm BT^2}/\phi_{\rm new}^{\rm BT^2}$ & 48.64 & 62.37 \\
\cline{2-5}
 & \multirow{2}{*}{OCA} & $\phi_{\rm new}^{\rm OCA}/\phi_{\rm old}$ & 26.19 & 40.65 \\
 &  & $\phi_{\rm new}^{\rm OCA}/\phi_{\rm new}^{\rm OCA}$ & 52.12 & 61.82 \\
\bottomrule[1pt]
\end{tabular}
\end{table}

\begin{table}[t]
\centering
\caption{Ablation study on the effect of orthogonality in $T_{\scriptscriptstyle{\pmb{\perp}}}$ and the cosine loss $\mathcal{L}_{\angle}$ in our method within a compatible learning setting. Results of the mean Average Precision (mAP) and Cumulative Match Characteristic (CMC) metrics are obtained from trials conducted on the CIFAR-50 and CIFAR-100 datasets. These experiments utilized the Resnet50-128 architecture for both the old and new models.}
\label{table:ablation2}
\setlength{\tabcolsep}{8pt}
\begin{tabular}{lccc}
\toprule[1pt]
\textbf{Method} & \textbf{Case} & \textbf{mAP@1.0} & \textbf{CMC-1} \\
\toprule[1pt]
Initial Model & $\phi_{\rm old}/\phi_{\rm old}$ & 23.32 &  31.32 \\
\hline
\multirow{2}{*}{Independent} & $\phi_{\rm new}^{\rm I}/\phi_{\rm old}$ & 01.29 & 01.02 \\
 & $\phi_{\rm new}^{\rm I}/\phi_{\rm new}^{\rm I}$ & 45.35 & 56.75 \\
\hline
\multirow{2}{*}{OCA w/o $T_{\scriptscriptstyle{\pmb{\perp}}}$, w/o $\mathcal{L}_{\angle}$} & $\phi_{\rm new}^{\rm OCA}/\phi_{\rm old}$ & 22.76 & 40.11 \\
 & $\phi_{\rm new}^{\rm OCA}/\phi_{\rm new}^{\rm OCA}$ & 48.54 & 59.78 \\
\hline
\multirow{2}{*}{OCA w/o $T_{\scriptscriptstyle{\pmb{\perp}}}$} & $\phi_{\rm new}^{\rm OCA}/\phi_{\rm old}$ & 25.89 & 40.11 \\
 & $\phi_{\rm new}^{\rm OCA}/\phi_{\rm new}^{\rm OCA}$ & 50.23 & 61.18 \\
\hline
\multirow{2}{*}{OCA w/o $\mathcal{L}_{\angle}$} & $\phi_{\rm new}^{\rm OCA}/\phi_{\rm old}$ & 26.13 & 40.34 \\
 & $\phi_{\rm new}^{\rm OCA}/\phi_{\rm new}^{\rm OCA}$ & 51.06 & 61.90 \\
\hline
\multirow{2}{*}{OCA} & $\phi_{\rm new}^{\rm OCA}/\phi_{\rm old}$ & \textbf{26.35} & \textbf{41.37} \\
 & $\phi_{\rm new}^{\rm OCA}/\phi_{\rm new}^{\rm OCA}$ & \textbf{52.06} & \textbf{62.02} \\
\bottomrule[1pt]
\end{tabular}
\end{table}

In the following, we present ablation studies on the influence of the dimensionality of the extra space $h_{\rm e}$ and each component of our method on compatible training on the CIFAR-100 dataset. Table \ref{table:ablation1} shows the results of our method compared to the BT$^2$ strategy in handling the additional representation space $h_{\rm e}$. We observe that our method does not suffer from increased dimensionality of $h_{\rm e}$, achieving consistent results across different sizes. Furthermore, we notice that increasing the dimensionality of the extra space allows the model to assimilate more new information, thereby achieving better results on both metrics when using new representations for both query and gallery sets. Instead, BT$^2$ struggles to learn compatible representations as the size of $h_{\rm e}$ increases. This is related to their model's change of basis architecture and the cosine loss that tries to match the geometrical structure of an independently trained model. This demonstrates the stability of our method in managing the new extra space compared to BT$^2$.

We present in Table \ref{table:ablation2} the results of our method with each component turned on and off. The results show that adding the cosine distance loss $\mathcal{L}_{\angle}$ to the BCT head improves performance by directly inducing alignment of the newly learned representations with the old class representation prototypes. The orthogonality in the transformation $T{\scriptscriptstyle{\pmb{\perp}}}$ also helps the model avoid disruption of compatible representations and inject new knowledge into the extra space, compared to a linear layer without orthogonality. When used together, the orthogonality of $T_{\scriptscriptstyle{\pmb{\perp}}}$ and $\mathcal{L}_{\angle}$ lead to state-of-the-art performance on the CIFAR-100 dataset.

\section{Conclusion}
\label{sec:conclusion}
This paper introduced an approach to manage the challenges associated with updating models in visual retrieval systems, particularly the need for backward compatibility and the high costs of backfilling. 
By expanding the feature space and applying an orthogonal transformation, our method allows for the integration of new information while maintaining compatibility with older models.

Our approach has demonstrated the potential to reduce operational complexities and costs traditionally involved with model updates, particularly in large-scale image datasets. 
The effectiveness of this method was assessed using the CIFAR-100 and ImageNet-1k datasets, where it was found to maintain compatibility and improve accuracy compared to existing methods.
\\\\
\noindent
\textbf{Acknowledgment:}
This work was partially supported by the European Commission under European Horizon 2020 Programme, grant number 951911 - AI4Media.
\\

\bibliographystyle{splncs04}
\bibliography{main}

\begin{thebibliography}{10}
\providecommand{\url}[1]{\texttt{#1}}
\providecommand{\urlprefix}{URL }
\providecommand{\doi}[1]{https://doi.org/#1}

\bibitem{biondi2023cores}
Biondi, N., Pernici, F., Bruni, M., Del~Bimbo, A.: Cores: Compatible representations via stationarity. IEEE Transactions on Pattern Analysis and Machine Intelligence  (2023)

\bibitem{biondi2023cl2r}
Biondi, N., Pernici, F., Bruni, M., Mugnai, D., Bimbo, A.D.: Cl2r: Compatible lifelong learning representations. ACM Transactions on Multimedia Computing, Communications and Applications  \textbf{18}(2s),  1--22 (2023)

\bibitem{biondi2024stationary}
Biondi, N., Pernici, F., Ricci, S., Del~Bimbo, A.: Stationary representations: Optimally approximating compatibility and implications for improved model replacements. In: Proceedings of the IEEE/CVF Conference on Computer Vision and Pattern Recognition (CVPR) (2024)

\bibitem{budnik2020asymmetric}
Budnik, M., Avrithis, Y.: Asymmetric metric learning for knowledge transfer. CVPR  (2021)

\bibitem{Chen_2019_CVPR}
Chen, K., Wu, Y., Qin, H., Liang, D., Liu, X., Yan, J.: {R3} adversarial network for cross model face recognition. In: {CVPR}. pp. 9868--9876. Computer Vision Foundation / {IEEE} (2019)

\bibitem{Cui_2024_CVPR}
Cui, Z., Zhou, J., Wang, X., Zhu, M., Peng, Y.: Learning continual compatible representation for re-indexing free lifelong person re-identification. In: Proceedings of the IEEE/CVF Conference on Computer Vision and Pattern Recognition (CVPR). pp. 16614--16623 (June 2024)

\bibitem{deng2009ImageNet}
Deng, J., Dong, W., Socher, R., Li, L.J., Li, K., Fei-Fei, L.: Imagenet: A large-scale hierarchical image database. In: 2009 IEEE conference on computer vision and pattern recognition. pp. 248--255. Ieee (2009)

\bibitem{duggal2021compatibility}
Duggal, R., Zhou, H., Yang, S., Xiong, Y., Xia, W., Tu, Z., Soatto, S.: Compatibility-aware heterogeneous visual search. In: Proceedings of the IEEE/CVF Conference on Computer Vision and Pattern Recognition. pp. 10723--10732 (2021)

\bibitem{iscen2020memory}
Iscen, A., Zhang, J., Lazebnik, S., Schmid, C.: Memory-efficient incremental learning through feature adaptation. In: European Conference on Computer Vision. pp. 699--715. Springer (2020)

\bibitem{Krizhevsky2009LearningML}
Krizhevsky, A.: {Learning Multiple Layers of Features from Tiny Images}. Technical report, Univ. Toronto (2009)

\bibitem{mccloskey1989catastrophic}
McCloskey, M., Cohen, N.J.: Catastrophic interference in connectionist networks: The sequential learning problem. In: Psychology of learning and motivation, vol.~24, pp. 109--165. Elsevier (1989)

\bibitem{meng2021learning}
Meng, Q., Zhang, C., Xu, X., Zhou, F.: Learning compatible embeddings. In: Proceedings of the IEEE/CVF International Conference on Computer Vision. pp. 9939--9948 (2021)

\bibitem{pernici2021regular}
Pernici, F., Bruni, M., Baecchi, C., Del~Bimbo, A.: Regular polytope networks. IEEE Transactions on Neural Networks and Learning Systems  (2021)

\bibitem{ramanujan2022forward}
Ramanujan, V., Vasu, P.K.A., Farhadi, A., Tuzel, O., Pouransari, H.: Forward compatible training for large-scale embedding retrieval systems. In: Proceedings of the IEEE/CVF Conference on Computer Vision and Pattern Recognition. pp. 19386--19395 (2022)

\bibitem{ramanujan2021forward}
Ramanujan, V., Vasu, P.K.A., Farhadi, A., Tuzel, O., Pouransari, H.: Forward compatible training for representation learning. Proceedings of the IEEE/CVF Conference on Computer Vision and Pattern Recognition  (2022)

\bibitem{robins1993catastrophic}
Robins, A.: Catastrophic forgetting in neural networks: the role of rehearsal mechanisms. In: Proceedings 1993 The First New Zealand International Two-Stream Conference on Artificial Neural Networks and Expert Systems. pp. 65--68. IEEE (1993)

\bibitem{seo2023online}
Seo, S., Uzunbas, M.G., Han, B., Cao, S., Zhang, J., Tian, T., Lim, S.N.: Online backfilling with no regret for large-scale image retrieval. arXiv preprint arXiv:2301.03767  (2023)

\bibitem{shen2020towards}
Shen, Y., Xiong, Y., Xia, W., Soatto, S.: Towards backward-compatible representation learning. In: Proceedings of the IEEE/CVF Conference on Computer Vision and Pattern Recognition. pp. 6368--6377 (2020)

\bibitem{Wan_2022_CVPR}
Wan, T.S.T., Chen, J.C., Wu, T.Y., Chen, C.S.: Continual learning for visual search with backward consistent feature embedding. In: Proceedings of the IEEE/CVF Conference on Computer Vision and Pattern Recognition (CVPR). pp. 16702--16711 (June 2022)

\bibitem{wang2020unified}
Wang, C., Chang, Y., Yang, S., Chen, D., Lai, S.: Unified representation learning for cross model compatibility. In: 31st British Machine Vision Conference 2020, {BMVC} 2020. {BMVA} Press (2020)

\bibitem{yan2021positive}
Yan, S., Xiong, Y., Kundu, K., Yang, S., Deng, S., Wang, M., Xia, W., Soatto, S.: Positive-congruent training: Towards regression-free model updates. In: Proceedings of the IEEE/CVF Conference on Computer Vision and Pattern Recognition. pp. 14299--14308 (2021)

\bibitem{yang2022we}
Yang, Y., Chen, S., Li, X., Xie, L., Lin, Z., Tao, D.: Inducing neural collapse in imbalanced learning: Do we really need a learnable classifier at the end of deep neural network? Advances in Neural Information Processing Systems  \textbf{35},  37991--38002 (2022)

\bibitem{zhang2021hot}
Zhang, B., Ge, Y., Shen, Y., Li, Y., Yuan, C., XU, X., Wang, Y., Shan, Y.: Hot-refresh model upgrades with regression-free compatible training in image retrieval. In: International Conference on Learning Representations (2021)

\bibitem{zhang2022towards}
Zhang, B., Ge, Y., Shen, Y., Su, S., Yuan, C., Xu, X., Wang, Y., Shan, Y.: Towards universal backward-compatible representation learning. arXiv preprint arXiv:2203.01583  (2022)

\bibitem{zhou2023bt}
Zhou, Y., Li, Z., Shrivastava, A., Zhao, H., Torralba, A., Tian, T., Lim, S.N.: Bt\^{} 2: Backward-compatible training with basis transformation. In: Proceedings of the IEEE/CVF International Conference on Computer Vision. pp. 11229--11238 (2023)

\end{thebibliography}
\end{document}